\documentclass[10pt, a4paper]{article}

\usepackage[final]{lrec2026} 

\usepackage{polyglossia}
\usepackage{amssymb}
\setmainlanguage{english}
\usepackage{afterpage}
\setotherlanguage{farsi}
\usepackage{pifont}
\newcommand{\xmark}{\ding{55}}

\usepackage{svg}            
\svgsetup{clean=true}       

\usepackage{xurl} 
\expandafter\def\expandafter\UrlBreaks\expandafter{\UrlBreaks%
  \do\a\do\b\do\c\do\d\do\e\do\f\do\g\do\h\do\i\do\j%
  \do\k\do\l\do\m\do\n\do\o\do\p\do\q\do\r\do\s\do\t%
  \do\u\do\v\do\w\do\x\do\y\do\z\do\A\do\B\do\C\do\D%
  \do\E\do\F\do\G\do\H\do\I\do\J\do\K\do\L\do\M\do\N%
  \do\O\do\P\do\Q\do\R\do\S\do\T\do\U\do\V\do\W\do\X%
  \do\Y\do\Z\do\0\do\1\do\2\do\3\do\4\do\5\do\6\do\7%
  \do\8\do\9\do\-}

\usepackage{multirow} 
\usepackage{amsmath}
\usepackage{listings}
\usepackage{xcolor}
\usepackage{booktabs,caption} 
\usepackage{mdframed}
\usepackage{fontspec}

\definecolor{boxbg}{HTML}{F7F5FF}
\definecolor{boxborder}{HTML}{5B4FCF}
\definecolor{promptbg}{HTML}{EEF2FF}
\definecolor{promptborder}{HTML}{7B6EE0}
\definecolor{answerbg}{HTML}{F0FFF4}
\definecolor{answerborder}{HTML}{38A169}
\definecolor{labelgray}{HTML}{6B7280}
\definecolor{goldtag}{HTML}{B7791F}

\newfontfamily\arabicfont[Script=Arabic]{Amiri-Regular.ttf}
\newfontfamily\arabicfontsf[Script=Arabic]{Amiri-Regular.ttf}

\usepackage{bidi}

\usepackage{soul}
\usepackage{xcolor} 

\title{TARAZ: Persian Short-Answer Question Benchmark for Cultural Evaluation of Language Models 
}

\name{\small Reihaneh Iranmanesh*\thanks{*Equal contribution.}, Saeedeh Davoudi*,
Pasha Abrishamchian, Ophir Frieder, Nazli Goharian} 

\address{\small IR Lab, Computer Science Department \\
         \small Georgetown University, Washington D.C. \\
    \small \{rei, saeedeh, pasha, ophir, nazli\}@ir.cs.georgetown.edu}

\abstract{
This paper presents a comprehensive evaluation framework for assessing the cultural competence of large language models (LLMs) in Persian. Existing Persian cultural benchmarks rely predominantly on multiple-choice formats and English-centric metrics that fail to capture Persian's morphological complexity and semantic nuance. Our framework introduces a Persian-specific short-answer evaluation that combines rule-based morphological normalization with a hybrid syntactic and semantic similarity module, enabling robust soft-match scoring beyond exact string overlap. Through systematic evaluation of 15 state-of-the-art open- and closed-source models across three culturally grounded Persian datasets, we demonstrate that our hybrid evaluation improves scoring consistency by +10\% compared to exact-match baselines by capturing meaning that surface-level methods cannot detect. Our human evaluation further confirms that the proposed semantic similarity metric achieves higher agreement with human judgments than LLM-based judges. We publicly release our evaluation framework, providing the first standardized benchmark for measuring cultural understanding in Persian and establishing a reproducible foundation for cross-cultural LLM evaluation research.
\\ \newline  }

\begin{document}

\maketitleabstract

\section{Introduction}

Large Language Models (LLMs) excel at general linguistic tasks~\cite{brown2020language, openai2025gpt5}, yet struggle with cultural common sense understanding ~\cite{shenculturalcommonsens}. Despite growing interest in cross-cultural evaluation~\cite{arora2025calmqa,li2024culturepark,li2024culturellm}, Persian remains notably absent from these major cultural benchmarks.

Few recent studies on the Persian language \cite{saffari2025iranian, moosavi2025percul} have explored incorporating cultural understanding in Multiple Choice Questions (MCQs) but recent research ~\cite{horbach2024llms} has shown that MCQ task assesses factual recall rather than conceptual understanding. Short-Answer Question (SAQ) tasks require free-form responses, providing a more direct probe of cultural and semantic understanding ~\cite{saffari2025iranian}. However, SAQ datasets and evaluation benchmarks that explicitly account for Persian cultural understanding remain limited. 

BLEnD~\cite{myung2024blend} is the first benchmark on SAQ task that is multilingual covering 13 languages including Persian (known as Farsi), providing opportunity for further research. Meanwhile, BLEnD suffers from culturally misaligned examples \cite{moosavi2025percul}. Their exact match evaluation metric cannot capture morphological Persian variations that express the same semantic content. For example, \textfarsi{نان} ("bread") versus \textfarsi{نون} ("bread," informal) are semantically identical but syntactically different. While some works employ LLM-based graders as an alternative, research demonstrates that these models exhibit instability, prompt sensitivity, and positional bias. ~\cite{luo2023,goyal2025ember,roitman2024exam}.

To address these challenges, we present TARAZ, a comprehensive evaluation framework that assesses LLMs' cultural competence in Persian through short-answer questions task.   
Our contributions are:\footnote{Code, data, and evaluation are available at \href{https://github.com/Georgetown-IR-Lab/TARAZ/}{TARAZ}.}

\begin{itemize}
\item ISN-SAQ and PerCul-SAQ: SAQ adaptation of the Iranian Social Norms (ISN) classification dataset~\cite{saffari2025iranian} and PerCul MCQ dataset \cite{moosavi2025percul}, transforming 1127 examples into culture-specific questions that test normative social behaviors across diverse environmental and demographic contexts.

\item Persian-specific LLMs' responses postprocessing, handling morphological and numeric variation (textual, and symbolic forms across Persian/Arabic numeral systems) and conjunction-based segmentation.

\item A hybrid syntactic and semantic-based evaluation to achieve robust semantic matching instead of exact matching. Our evaluation framework is extensible to new datasets and any language model's responses. 

\item Comprehensive evaluation of 15 state-of-the-art (multilingual and Persian) LLMs across six metrics on BLEnD, ISN-SAQ and PerCul-SAQ datasets. 

\end{itemize}

In the rest of this paper, we first review the background and related works in 
Section~\ref{sec:background}. Section~\ref{sec:datasets} presents the datasets used in our study, 
followed by Section~\ref{sec:models}, which describes the models employed. Section ~\ref{sec:evaluation} details our evaluation framework, and 
Section~\ref{sec:experiments} reports the experimental setup and results. 
Finally, Section~\ref{sec:conclusion} concludes the paper.

\section{Related Work}
\label{sec:background}

In the following section, we provide a review of studies related to the evaluation of LLMs in Persian and cultural settings.

\subsection{Persian NLP Benchmarks and Evaluation Challenges}

Persian NLP has advanced through development of diverse benchmarks. ParsiNLU~\cite{khashabi2021parsinlu} provided early task coverage with 2.4K multiple-choice questions across literature, commonsense, and mathematics domains. ParsBench~\cite{shariati2025parsbench} established standardized evaluation protocols and developed a leaderboard ranking Persian LLMs on tasks including sentiment analysis, machine translation, and multiple-choice question answering. The Khayyam Challenge~\cite{ghahroodi2024khayyam} adapted MMLU with 20,192 four-choice questions across 38 domains extracted from Persian examinations, introducing content spanning elementary to secondary education levels. These efforts employ MCQ formats, which measure factual recall and perform exact matching rather than comprehension~\cite{luo2023}.

Hosseinbeigi et al.~\cite{hosseinbeigi2025advancing}\footnote{Their dataset and evaluation code are not publicly available as of this submission.} introduced PeKA (3,600 MCQs on Persian cultural knowledge) and PK-BETS (4,000 questions including open-ended generation tasks). However, their open-ended evaluation relies on LLM-as-a-Judge (GPT-4o), achieving only moderate agreement with human judgment (Cohen's $\kappa$=0.54), introducing evaluation bias, proprietary model dependency, and scalability limitations.

FarsEval-PKBETS~\cite{shamsfard2025farseval}\footnotemark[\value{footnote}] introduced 3500 multiple-choice, and 500 short-answer questions covering medicine, law, religion, and Persian language tasks. However, their evaluation of short-answer questions relies on manual human annotation with subjective Correct/Wrong/Semi-correct labels, limiting scalability and potentially introducing inconsistency. They evaluated only three models and provided no automated evaluation framework for handling morphological variation in Persian SAQs. 
BLEnD~\cite{myung2024blend} is introduced as the first open-source SAQ and MCQ benchmark that includes Farsi. It provides both datasets and evaluation metrics for the SAQ task. The BLEnD dataset is discussed in detail in Section~\ref{sec:datasets}.

\subsection{Multilingual and Cultural-Linguistic Alignment}

Ying et al.\cite{ying2025disentangling} showed that multilingual models perform better when the question’s language aligns with its cultural context. CulFiT~\cite{zhang2025culfit} introduces multilingual critique data synthesis with fine-grained rewards, achieving state-of-the-art open-source performance but still faces SAQ evaluation challenges.
Evaluating models only in English uses translation-based methods that do not show real understanding of the target language. For example, cultural concepts like Taarof, a kind of polite social ritual in Farsi, have no clear English equivalent.
TaarofBench \cite{sadr2025taarofbench} confirms this pattern with dramatic improvements from English to Persian prompts: DeepSeek V3 improved by 32.0 points, GPT-4o by 33.1 points, and Claude 3.5 by 25.2 points. Matina ~\cite{matina2025} introduces a multi-expert approach based on Low-Rank Adaptation (LoRA) to better align Persian text generation with cultural and linguistic nuances\footnote{Their model is not publicly available as of this submission.}. 

Our Persian-specific postprocessing (numeric normalization, conjunction segmentation, morphological normalization) addresses the evaluation gaps that prior work~\cite{shamsfard2025farseval,zhang2025culfit, myung2024blend} could not resolve through manual effort alone, providing an automated and scalable solution for robust Persian SAQ evaluation.

\subsection{LLM-Based Evaluation and Its Limitations}
Chamieh et al.\cite{horbach2024llms} demonstrate that zero-shot and few-shot LLM graders underperform supervised baselines. Luo et al.\cite{luo2023} show that even calibrated rubric prompting achieves only modest human agreement, typically plateauing at 70--80\text{\%} regardless of prompt engineering effort. The EMBER benchmark~\cite{goyal2025ember} reveals that LLM judges systematically penalize epistemic uncertainty markers, reducing scoring fairness for cautious but correct responses. The EXAM++ framework~\cite{roitman2024exam} documents technical pitfalls including positional biases (15--20\% preference for first-shown answers) and brittle answer verification where minor formatting differences cause evaluation failure.
Traditional SAQ approaches ~\cite{myung2024blend} rely on string-matching metrics (exact match, soft match) which fail to recognize semantically equivalent but syntactically distinct answers--a problem exacerbated in morphologically rich languages like Persian. 
\section{Datasets}
\label{sec:datasets}

We evaluate LLMs on three Persian cultural datasets, each targeting distinct aspects of cultural understanding. All datasets use native Persian prompts and references to ensure authentic evaluation without translation artifacts.

\paragraph{BLEnD} 
The BLEnD (Benchmark for LLMs on Everyday Knowledge in Diverse Cultures and Languages)~\cite{myung2024blend} dataset provides everyday cultural questions across 13 languages and 16 regions. We utilize the Persian data containing 500 questions spanning topics like food, sports, family, education, holidays/celebrations/leisure, and work-life. Each question has multiple human-annotated acceptable answers with associated frequency counts, enabling weighted evaluation. Questions are designed to probe implicit cultural knowledge that locals take for granted but may be absent from multilingual training corpora. We exclude questions where annotators predominantly responded with "I don't know" (idks $\geq$ 5) or "not applicable" responses ($\geq$ 3), following the original BLEnD paper. See Figure~\ref{fig:blend-example} for a culturally-specific example.

\paragraph{PerCul-SAQ} 
The PerCul (Persian Cultural Understanding)~\cite{moosavi2025percul} dataset evaluates story-driven cultural reasoning through 592 multiple-choice questions derived from Persian cultural scenarios. Each entry consists of a Persian narrative describing a cultural situation, followed by a question with four answer choices. We introduce PerCul-SAQ, converting PerCul to short-answer format by changing the prompt, creating a more challenging generation task than the original multiple-choice setting. Stories contextualize cultural elements within realistic scenarios, requiring models to infer appropriate behavior or identify culturally-significant items rather than merely recalling facts.

The dataset spans 11 cultural categories: foods (\textfarsi{خوراکی}، 191 questions), visible behavior (\textfarsi{رفتار}، 56), iconic figures (\textfarsi{چهره نمادین}، 55), institutions (\textfarsi{نهاد}، 43), architecture (\textfarsi{معماری}، 43), objects (\textfarsi{شی}، 42), appropriacy (\textfarsi{هنجار}، 36), dress (\textfarsi{پوشاک}، 33), music (\textfarsi{موسیقی/ساز}، 32), art (\textfarsi{هنر}، 32), and rituals (\textfarsi{آیین}، 29). 

See Figure~\ref{fig:percul-example} for an example where a Nowruz celebration story requires identifying the Tombak (\textfarsi{تنبک}), 
a single-headed goblet drum central to Persian musical tradition.

\paragraph{ISN-SAQ}
We derive our ISN-SAQ dataset from the Persian Social Norms (PSN) dataset~\cite{saffari2025iranian}, which contains 1,699 Iranian social norms with environment, context, and cultural appropriateness labels (Expected/Normal/Taboo). We convert norms labeled as Expected or Normal into short-answer questions, filtering out general norms and duplicates, yielding 535 culture-specific questions.

The questions cover diverse social situations: airport etiquette (e.g., \textfarsi{در فرودگاه‌های ایران قبل از سفر چه رفتاری رایج است؟} / "What behavior is common at Iranian airports before travel?" $\rightarrow$ \textfarsi{طلب حلالیت} / requesting forgiveness), marketplace practices (\textfarsi{در بازار ایران هنگام خرید با فروشندگان چه رفتاری رایج است؟} $\rightarrow$ \textfarsi{چانه‌زنی} / bargaining), family traditions (\textfarsi{بعد از تولد پسران در ایران چه مراسمی برگزار می‌شود؟} $\rightarrow$ \textfarsi{ختنه سوران} / circumcision ceremony), and gender-specific behaviors. ISN-SAQ emphasizes situated social norms tied to specific environments (\textfarsi{مسجد}/mosque, \textfarsi{بازار}/marketplace, \textfarsi{فرودگاه}/airport, \textfarsi{خانه}/home) and demographic contexts, reflecting how appropriate behavior varies across settings within Iranian culture.

Figure~\ref{fig:psn-example} illustrates how a Bazaar norm 
(\textfarsi{چانه‌زنی}, bargaining) is reformulated as an SAQ, 
capturing context-dependent social behavior specific to Iranian 
marketplace culture.

Table~\ref{tab:dataset_stats} summarizes dataset characteristics. BLEnD emphasizes everyday factual knowledge with multiple valid answers per question, PerCul-SAQ tests cultural narrative reasoning with single correct answers, and ISN-SAQ focuses on normative social behaviors with highly specific contextual requirements. BLEnD, PerCul-SAQ, and ISN-SAQ each target a single facet of Persian 
cultural competence -- everyday facts, story-based reasoning, and social norms  respectively. No prior work evaluates all three jointly or provides a 
unified SAQ scoring framework applicable across them. Our work, TARAZ, fills this gap.

\begin{table}[t]
\centering
\resizebox{\columnwidth}{!}{%
\begin{tabular}{@{}lccc@{}}
\toprule
& \textbf{BLEnD} & \textbf{PerCul-SAQ} & \textbf{ISN-SAQ} \\
\midrule
Questions & 500 & 592 & 535 \\
Answers/Q & Multi (3.7) & Single & Single \\
Avg. Q len. & 11.0 & 99.5 & 9.1 \\
Avg. A len. & 1.6 & 3.1 & 2.7 \\
Focus & Knowledge & Stories & Norms \\
\bottomrule
\end{tabular}%
}
\caption{
Comparison of short-answer question (SAQ) datasets. 
\textit{Questions} indicates the total number of questions; 
\textit{Answers/Q} shows the number of valid answers per question; 
\textit{Avg.~Q len.} and \textit{Avg.~A len.} denote average question and answer lengths in words (whitespace-tokenized); 
and \textit{Focus} describes each dataset's main content type. 
Note that PerCul-SAQ questions are narrative scenarios rather than direct questions, explaining the longer average length.
}
\label{tab:dataset_stats}

\end{table} 

 \section{Models} 
\label{sec:models}
We evaluate 15 generative models spanning three categories: closed-source proprietary models, open-weight foundation models, and Persian fine-tuned models. Since SAQ evaluation requires text generation, we focus on Generative Language Models and exclude encoder-based models (e.g., ParsBERT~\cite{farahani2021parsbert}, which are designed for classification rather than open-ended response generation). Persian models were selected based on their ranking across three major Persian LLM leaderboards: MIZAN ~\cite{mizan}, ParsBench ~\cite{shariati2025parsbench}, and Open Persian LLM Leaderboard ~\cite{PartAI_OpenPersianLLM_2024}.

\subsection{Closed Source Models}
We evaluate four frontier proprietary models with API-based access: \textit{GPT-5}~\cite{openai2025gpt5}, \textit{Claude Opus 4.1} ~\cite{anthropic2025claude4}, \textit{Claude Sonnet 4.5}~\cite{anthropic2025sonnet45}, and \textit{Gemini 2.5 Flash-Lite}~\cite{google2025gemini25}.

\subsection{Open Weight Models}
We evaluate five open-weight foundation models: \textit{Qwen2.5-72B}~\cite{qwen2025qwen25} (multilingual with explicit Persian support), \textit{LLaMA 3.1-70B}~\cite{touvron2023llama}, \textit{DeepSeek-V3}~\cite{deepseek2024} (mixture-of-experts architecture), \textit{Gemma-2-27B}~\cite{team2025gemma}, and \textit{Phi-4-14B}~\cite{microsoft2025phi4}.

\subsection{Persian Fine-Tuned Models}
We evaluate six Persian-adapted models representing diverse adaptation strategies: \textit{PersianMind}~\cite{rostami2024persianmind} extends LLaMA2-7B with 10,000 Persian tokens and trains on 2B Persian tokens; \textit{ParsT5}~\cite{parst5} adapts T5 encoder-decoder through masked language modeling on 35GB Persian text; \textit{Dorna}~\cite{dorna2024} fine-tunes LLaMA 3-8B-Instruct on Persian instructions; \textit{PersianLLaMA}~\cite{abbasi2023persianllama} (13B) trains adapters on pretrained LLaMA2; \textit{Maral}~\cite{maralgpt2023} fine-tunes Mistral-7B on machine-translated Alpaca Persian; and \textit{AVA-LLAMA}~\cite{moghadam2024ava} applies parameter-efficient fine-tuning to LLaMA 3-8B.

\section{Evaluation}
\label{sec:evaluation}

\subsection{Evaluation Metrics}
To evaluate the capabilities of different language models on Persian cultural knowledge, we use syntactic metrics (Exact Match and ROUGE-1, ROUGE-2, ROUGE-L \cite{lin2004rouge}), semantic-based approaches including GPT-5 as a judge, and cosine similarities using Transformer-based Persian embeddings. All metrics assess responses of LLMs against ground truth data. At the end, all techniques assign a binary score to each question–answer pair: “1” if the answer is correct and “0” otherwise. The definitions of each metric are provided below:
\hspace{0.8em}
\begin{enumerate}
\item \textbf{Exact Match (EM)}: Normalized Responses are lemmatized using the BLEnD benchmark. A response is labeled as "1" as long as it exactly matches one of the ground-truth data.

\item \textbf{ROUGE Metrics:} We employ three ROUGE metrics~\cite{lin2004rouge}: (1) \textit{ROUGE-1} measures unigram overlap between the model response and ground truth; (2) \textit{ROUGE-2} measures bigram overlap (consecutive word pairs); and (3) \textit{ROUGE-L} calculates the longest common subsequence. All scores are computed as F1 measures. If the F1 score is $1.0$, we label the response as "1"; otherwise, it will be labeled as "0".

\item \textbf{LLM-judge:} GPT-5 evaluates semantic equivalence of model responses against ground truth answers using the following prompt:

\begin{lstlisting}{python}
text = f"""
You are evaluating a model's answer.

Question ID: {qid}
Response: "{response}"

Valid answers: {answers}

Does the response semantically match 
ANY of the valid answers?
Answer ONLY with 1 if it matches, 
0 if it does not.
"""
\end{lstlisting} 
If GPT-5 returns 1, the answer is labeled as ``1''; otherwise, it is labeled as ``0.''

\item \textbf{Cosine similarity using a Transformer-based model.} \textit{maux-gte-persian-v3} \cite{mauxgte2024} represents a high-performance Persian sentence embedding solution fine-tuned from Alibaba-NLP/gte-multilingual-base \cite{zhang-etal-2024-mgte}. This model produces 768-dimensional sentence embeddings with support for sequences up to 8192 tokens, making it suitable for long-form Persian text processing. Responses with cosine similarity above an empirically determined threshold $0.85$ are classified as "1", and "0" otherwise.
Using this model, we are able to generate direct sentence-level embeddings, allowing for more sophisticated semantic understanding. The system calculates the maximum sentence similarity between any sentence in the model response and any ground truth annotations.

\item \textbf{Hybrid approach:} We applied Persian-specific postprocessing (Section~\ref{subsec:postprocessing}) on each model's responses before computing cosine similarity using \textit{maux-gte-persian-v3} embeddings. In that way, we could handle numeric normalization (e.g., \textfarsi{۱۲۳} vs \textfarsi{صد و بیست و سه}), conjunction-based segmentation, morphological normalization, and diacritic removal that were not handled by previous benchmarks. The hybrid method is the most robust metric for Persian evaluation (Section \ref{subsec:postprocessing}).
\end{enumerate}

\subsection{Postprocessing}
\label{subsec:postprocessing}
Persian presents persistent challenges for natural language processing due to its morphological complexity, orthographic ambiguity, and inconsistent writing conventions. The absence of clear word boundaries, where clitics and compound words are often written without spaces, complicates tokenization and lemmatization. Orthographic inconsistencies, such as optional half-spaces and variable diacritic use, result in multiple valid spellings of the same word, for example, \textfarsi{می رود} ("goes") versus \textfarsi{میرود} ("goes"). Similarly, lexical variation between formal and colloquial registers creates normalization challenges, as in \textfarsi{نان} ("bread") versus \textfarsi{نون} ("bread," informal). Moreover, Persian text often mixes symbols, numbers, and characters from different writing systems including multiple numeral encodings (e.g., \textfarsi{۶} vs. 6, \textfarsi{ششم} vs \textfarsi{شش} vs \textfarsi{شیش}), inconsistent use of Arabic and Persian letters (e.g., \textfarsi{یک} vs \textfarsi{یك}), and diverse representations of dates (e.g., \textfarsi{۱۴۰۰} / \textfarsi{۰۷} / \textfarsi{۲۵} → "the twenty-fifth of Mehr, fourteen hundred") and times (e.g., \textfarsi{۱۱:۳۵} → "eleven thirty-five"). These examples illustrate how surface variability and non-standardized writing conventions introduce substantial ambiguity, making Persian an especially demanding language for robust text normalization and evaluation.
During postprocessing, we address these issues to ensure consistent and comparable evaluation of Persian texts:
\begin{itemize}
    \item \textit{Whitespace and punctuation normalization:} Removes extra spaces and trailing punctuation marks such as "\textfarsi{.}", "\textfarsi{،}", "\textfarsi{؛}", "\textfarsi{!}", and "\textfarsi{؟}".
    \item \textit{Digit normalization:} Converts both English (0-9) and Arabic (\textfarsi{۰} - \textfarsi{۹}) numerals into their Persian equivalents (\textfarsi{۰} - \textfarsi{۹}) (e.g., 56 vs \textfarsi{۵۶} vs \textfarsi{٥٦}).
    \item \textit{Character normalization:} Replaces Arabic variants of certain letters with their Persian counterparts, for example:
    \begin{itemize}
        \item \textfarsi{ي} $\rightarrow$ \textfarsi{ی}
        \item \textfarsi{ك} $\rightarrow$ \textfarsi{ک}
        \item \textfarsi{ة} $\rightarrow$ \textfarsi{ه}
        \item Removes diacritics and redundant \textfarsi{ء}.
    \end{itemize}
    \item \textit{Diacritic removal:} Eliminates Persian vowel marks to standardize text.
    \item \textit{Stop word removal:} Removes common Persian stop words using the \texttt{hazm} library, keeping only meaningful tokens.
    \item \textit{Suffix removal (stemming):} Strips frequent Persian suffixes such as \textfarsi{ات}, \textfarsi{یات}, \textfarsi{ها}, \textfarsi{ان}, \textfarsi{ین}, \textfarsi{ون}, and \textfarsi{گان} to reduce words to their base form.
    \item Splits text using Persian and English punctuation, including semicolons (\textfarsi{؛}, \textit{;}) and commas (\textfarsi{،}, \textit{,}).
    \item Further splitting items by conjunctions when they occur \emph{between two substantial parts}:
    \begin{itemize}
        \item \textfarsi{و} (and)
        \item \textfarsi{یا} (or)
        \item \textfarsi{هم} (also)
        \item \textfarsi{همینطور} (likewise)
        \item \textfarsi{نیز} (also)
        \item \textfarsi{همچنین} (in addition)
    \end{itemize}
    \item Removes any empty or redundant entries, returning a list of clean, distinct response items.
\end{itemize}

\section{Experiments and Results}
\label{sec:experiments}

\begin{table*}[!ht] 
\centering
\resizebox{\textwidth}{!}{%
\begin{tabular}{@{}llcccccccc@{}}
\toprule
& \textbf{Model} & \textbf{EM} & \textbf{ROUGE-1} & \textbf{ROUGE-2} & \textbf{ROUGE-L} & \textbf{LLM-judge} & \textbf{Maux} & \textbf{Maux+Post} \\
\midrule
\multirow{4}{*}{\rotatebox[origin=c]{90}{\small Closed Source}} 
& GPT-5 & 73.246 & 42.105 & 12.939 & 42.105 & 71.930 & 84.652 & 85.837 \\
& Claude-Opus-4.1 & 79.386 & 44.737 & 11.184 & 44.739 & 81.579 & 84.869 & 85.781 \\
& Claude-Sonnet-4.5 & 74.561 & 42.982 & 9.868 & 42.763 & 78.509 & 85.307 & 86.764 \\
& Gemini-2.5-Flash-Lite & 54.605 & 34.649 & 11.404 & 21.711 & 58.772 & 62.061 & 62.854 \\
\addlinespace[0.5em]
\midrule
\addlinespace[0.5em]
\multirow{5}{*}{\rotatebox[origin=c]{90}{\small Open Weight}} 
& LLaMA-3.1-70B-Inst & 47.368 & 21.491 & 6.140 & 21.491 & 57.018 & 63.158 & 64.493 \\
& DeepSeek-V3 & 43.421 & 21.711 & 5.921 & 42.213 & 45.833 & 48.247 & 50.012 \\
& Gemma-2-27B-IT & 70.614 & 37.500 & 8.991 & 37.500 & 74.123 & 78.947 & 79.391 \\
& Phi-4-14B & 25.000 & 5.263 & 0.439 & 5.263 & 23.465 & 72.012 & 73.204 \\
& Qwen-2.5-72B-Instruct & 53.289 & 23.465 & 4.386 & 23.246 & 31.140 & 73.521 & 74.812 \\
\addlinespace[0.5em]
\midrule
\addlinespace[0.5em]
\multirow{6}{*}{\rotatebox[origin=c]{90}{\small Persian Fine-Tuned}} 
& PersianMind-v1.0-7B & 32.018 & 4.605 & 1.316 & 4.600 & 25.000 & 27.227 & 28.018 \\
& ParsT5 & 1.974 & 0.219 & 0.219 & 4.005 & 0.877 & 10.151 & 10.174 \\
& Dorna-LLaMA3-8B-Instruct & 49.123 & 20.395 & 3.070 & 20.395 & 47.807 & 54.824 & 55.064 \\
& PersianLLaMA-13B & 21.930 & 15.965 & 1.681 & 15.967 & 0.000 & 5.204 & 6.214 \\
& Maral-7B & 18.860 & 9.211 & 1.535 & 8.991 & 20.175 & 21.181 & 22.229 \\
& AVA-LLaMA-3-8B & 49.781 & 30.711 & 3.006 & 34.429 & 39.333 & 42.359 & 58.647 \\
\bottomrule
\end{tabular}%
}

\caption{Performance comparison of closed-source, open-weight, and Persian fine-tuned language models on the BLEnD dataset. Reported metrics include Exact Match (EM), ROUGE-1, ROUGE-2, and ROUGE-L for lexical similarity; LLM-judge, Maux and Maux+Post for semantic evaluation. Closed-source models (e.g., GPT-5, Claude) outperform others across most metrics, while open-weight models like Gemma-2 and fine-tuned Persian models (e.g., AVA-LLAMA3-8B) demonstrate competitive performance within their respective categories.}
\label{tab:BLEnD}
\end{table*}

\begin{table*}[!ht] 
\centering
\resizebox{\textwidth}{!}{%
\begin{tabular}{@{}llcccccccc@{}}
\toprule
& \textbf{Model} & \textbf{EM} & \textbf{ROUGE-1} & \textbf{ROUGE-2} & \textbf{ROUGE-L} & \textbf{LLM-judge} & \textbf{Maux} & \textbf{Maux+Post} \\
\midrule
\multirow{4}{*}{\rotatebox[origin=c]{90}{\small Closed Source}} 
& GPT-5 & 45.946 & 29.899 & 18.750 & 29.899 & 74.493 & 79.493 & 80.493 \\
& Claude-Opus-4.1 & 39.358 & 29.899 & 16.047 & 29.899 & 60.304 & 65.304 & 66.304 \\
& Claude-Sonnet-4.5 & 38.345 & 25.338 & 13.176 & 25.338 & 57.770 & 62.770 & 63.770 \\
& Gemini-2.5-Flash-Lite & 19.426 & 15.372 & 7.264 & 15.203 & 35.473 & 40.473 & 41.473 \\
\addlinespace[0.5em]
\midrule
\addlinespace[0.5em]
\multirow{5}{*}{\rotatebox[origin=c]{90}{\small Open Weight}} 
& LLaMA-3.1-70B-Inst & 12.162 & 7.264 & 2.703 & 7.095 & 21.622 & 26.622 & 27.622 \\
& DeepSeek-V3 & 30.912 & 23.986 & 11.824 & 23.818 & 48.649 & 53.649 & 54.649 \\
& Gemma-2-27B-IT & 17.905 & 14.696 & 8.277 & 14.527 & 30.236 & 35.236 & 36.236 \\
& Phi-4-14B & 1.182 & 0.000 & 0.000 & 0.000 & 4.223 & 9.223 & 10.223 \\
& Qwen-2.5-72B-Instruct & 3.716 & 2.365 & 1.182 & 2.365 & 17.230 & 22.230 & 23.230 \\
\addlinespace[0.5em]
\midrule
\addlinespace[0.5em]
\multirow{6}{*}{\rotatebox[origin=c]{90}{\small Persian Fine-Tuned}} 
& PersianMind-v1.0-7B & -- & 0.000 & 0.000 & 0.000 & 23.142 & 28.142 & 29.142 \\
& ParsT5 & 0.000 & 0.000 & 0.000 & 0.000 & 1.689 & 6.689 & 7.689 \\
& Dorna-LLaMA3-8B-Instruct & 7.284 & 0.845 & 0.189 & 0.845 & 9.459 & 14.459 & 15.459 \\
& PersianLLaMA-13B & 0.845 & 0.000 & 0.000 & 0.000 & 20.101 & 25.101 & 26.101 \\
& Maral-7B & -- & 0.000 & 0.000 & 0.000 & 33.446 & 38.446 & 39.446 \\
& AVA-LLaMA-3-8B & 5.068 & 0.000 & 0.000 & 0.000 & 6.588 & 11.588 & 12.588 \\
\bottomrule
\end{tabular}%
}

\caption{Performance of closed-source, open-weight, and Persian fine-tuned models on the PerCul-SAQ dataset, which evaluates story-based cultural reasoning across 11 cultural categories. Metrics include Exact Match (EM), ROUGE-1, ROUGE-2, and ROUGE-L for lexical overlap; LLM-judge for semantic equivalence; and Maux and Maux+Post for human-aligned cultural judgment. Closed-source models achieve higher overall scores, while fine-tuned Persian models show low capability in culturally grounded comprehension tasks.}
\label{tab:percul}
\end{table*}

\begin{table*}[!ht] 
\centering
\resizebox{\textwidth}{!}{%
\begin{tabular}{@{}llcccccccc@{}}
\toprule
& \textbf{Model} & \textbf{EM} & \textbf{ROUGE-1} & \textbf{ROUGE-2} & \textbf{ROUGE-L} & \textbf{LLM-judge} & \textbf{Maux} & \textbf{Maux+Post} \\
\midrule
\multirow{4}{*}{\rotatebox[origin=c]{90}{\small Closed Source}} 
& GPT-5 & 20.187 & 8.224 & 4.673 & 6.224 & 41.682 & 46.682 & 46.682 \\
& Claude-Opus-4.1 & 20.374 & 8.037 & 3.925 & 7.664 & 42.056 & 47.056 & 47.056 \\
& Claude-Sonnet-4.5 & 18.131 & 7.684 & 3.925 & 7.477 & 37.757 & 42.757 & 42.757 \\
& Gemini-2.5-Flash-Lite & 16.636 & 9.346 & 4.299 & 3.346 & 33.084 & 38.084 & 38.084 \\
\addlinespace[0.5em]
\midrule
\addlinespace[0.5em]
\multirow{5}{*}{\rotatebox[origin=c]{90}{\small Open Weight}} 
& LLaMA-3.1-70B-Inst & 11.589 & 6.729 & 3.364 & 6.729 & 27.290 & 32.290 & 32.209 \\
& DeepSeek-V3 & 17.944 & 4.120 & 1.301 & 4.120 & 44.299 & 49.299 & 49.229 \\
& Gemma-2-27B-IT & 13.271 & 6.729 & 2.991 & 6.729 & 32.150 & 37.150 & 37.150 \\
& Phi-4-14B & 3.364 & 0.374 & 0.000 & 0.374 & 9.346 & 14.346 & 14.346 \\
& Qwen-2.5-72B-Instruct & 9.346 & 1.495 & 0.935 & 1.495 & 31.776 & 36.776 & 36.764 \\
\addlinespace[0.5em]
\midrule
\addlinespace[0.5em]
\multirow{6}{*}{\rotatebox[origin=c]{90}{\small Persian Fine-Tuned}} 
& PersianMind-v1.0-7B & 3.178 & 2.056 & 1.308 & 2.056 & 8.224 & 14.224 & 14.224 \\
& ParsT5 & 0.187 & 0.000 & 0.000 & 0.000 & 0.561 & 5.561 & 6.561 \\
& Dorna-LLaMA3-8B-Instruct & 4.299 & 1.121 & 0.748 & 1.121 & 15.327 & 20.327 & 20.327 \\
& PersianLLaMA-13B & 2.617 & 0.000 & 0.000 & 0.000 & 5.981 & 10.981 & 10.981 \\
& Maral-7B & 0.374 & 0.000 & 0.000 & 0.000 & 1.869 & 6.869 & 6.869 \\
& AVA-LLaMA-3-8B & 4.486 & 1.121 & 0.748 & 1.121 & 15.701 & 20.701 & 20.701 \\
\bottomrule
\end{tabular}%
}
\caption{Model performance on the ISN-SAQ dataset across key metrics (EM, ROUGE, LLM-judge, and Maux scores). Maux+Post metric achieves the strongest results overall, with Claude Opus and DeepSeek-V3 leading. Persian fine-tuned models perform notably lower, the same pattern we observed in previous datasets.}
\label{tab:isn}
\end{table*}

Our experimental plan evaluates LLMs on their ability to understand and generate culturally grounded responses in Farsi using short-answer questions (SAQs). Evaluation is conducted on three datasets: BLEnD, PerCul-SAQ, ISN-SAQ, which together cover everyday culture, narratives, and normative social behaviors. Each model receives only Farsi prompts to ensure evaluation within native linguistic context. We report Exact Match, ROUGE-1, ROUGE-2, ROUGE-L, GPT-5 as LLM-judge, Maux, and Maux+Post metrics. All evaluations are performed in a zero-shot setting with fixed temperature (T = 0.2) if applicable, and outputs are normalized before scoring. Finally, we analyze results across cultural domains and model types to further understand the strengths and weaknesses of each model.

\paragraph{Overall Performance}

\begin{figure*}[!ht]
  \centering
  \includegraphics[width=\textwidth]{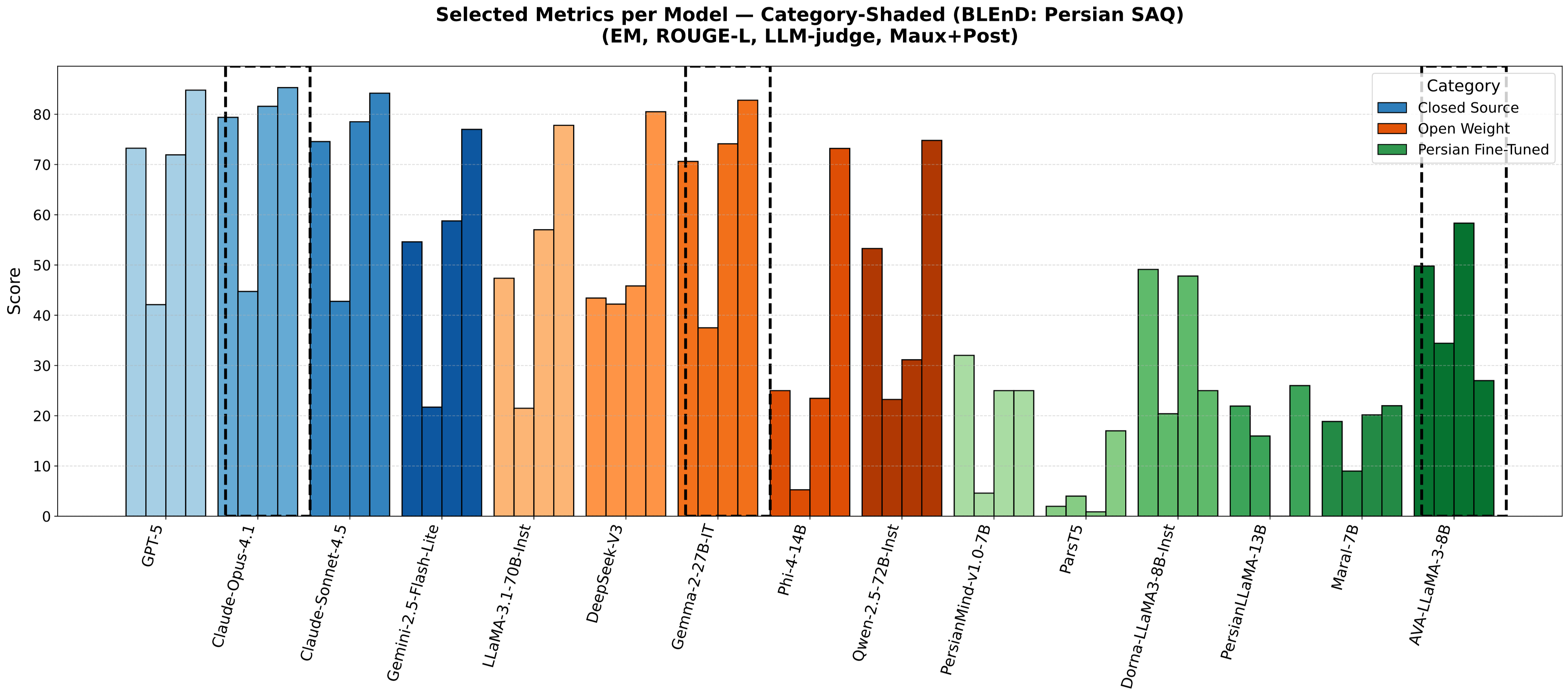}
  \caption{Comparison of model performance on the BLEnD dataset using four key metrics (left to right) -- Exact Match (EM), ROUGE-L, LLM-judge, and Maux+Post. Bars are grouped per model and color-coded by category (Closed Source, Open Weight, Persian Fine-Tuned). Each dashed box highlights the best-performing model within its category, based on normalized average performance across all selected metrics. Claude-Opus-4.1 is the best closed-source model, Gemma-2-27B-IT is the best open-weight model and AVA-LLaMA-3-8B is the best Persian fine-tuned model. Closed-source models achieve the highest scores overall.}
  \label{fig:blend-svg}
\end{figure*}


\begin{figure}[!ht]
  \centering
\includegraphics[width=\columnwidth]{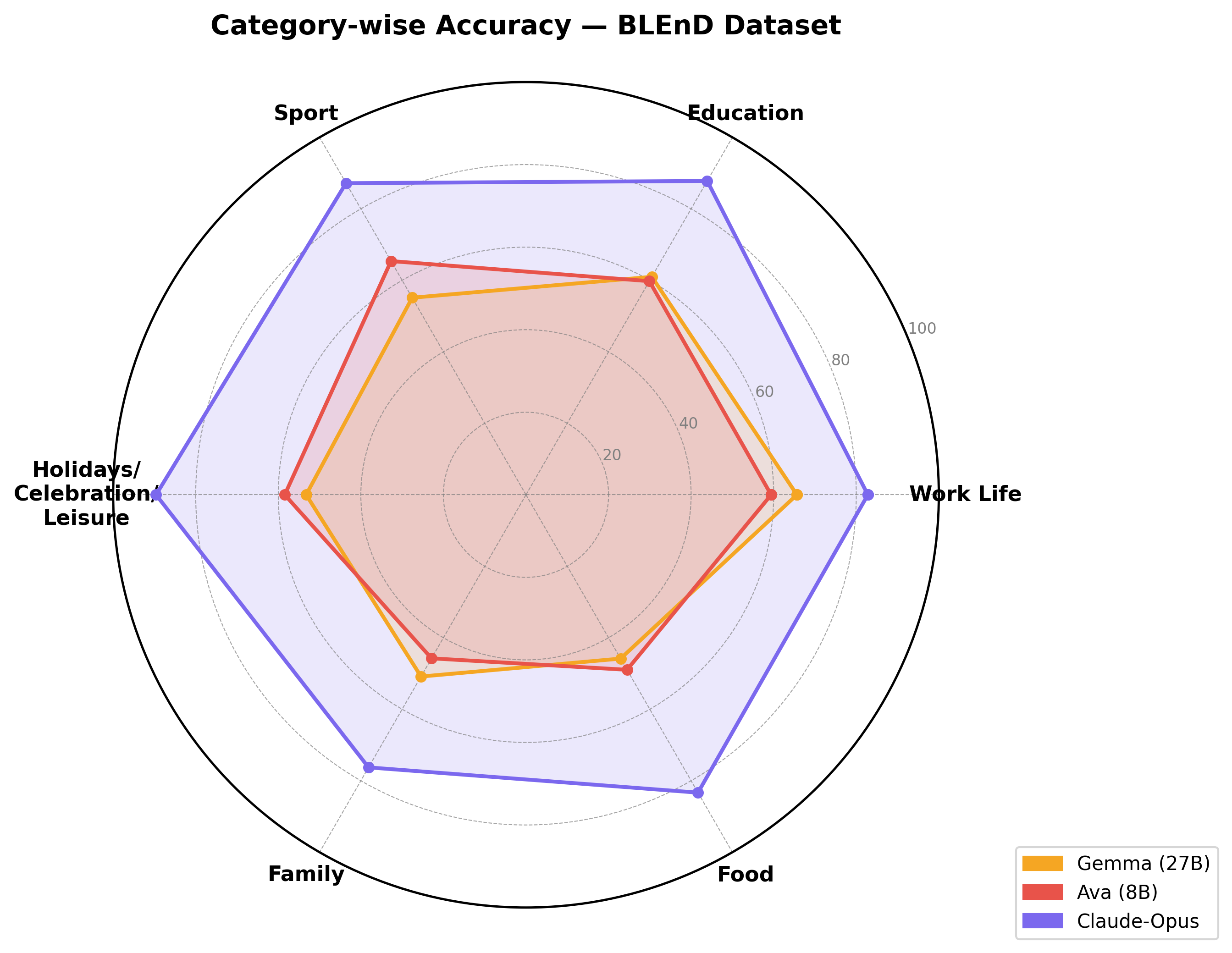}
  \caption{Category-wise accuracy on BLEnD. The plot shows accuracy for the top three models on BLEnD dataset. Claude Opus shows better accuracy among different categories compared to Gemma-2-27-IT and Ava-LLaMA-3-8B Persian model.}
  \label{fig:blend-spider}
\end{figure}

\begin{figure}[!ht]
  \centering
\includegraphics[width=0.5\textwidth]{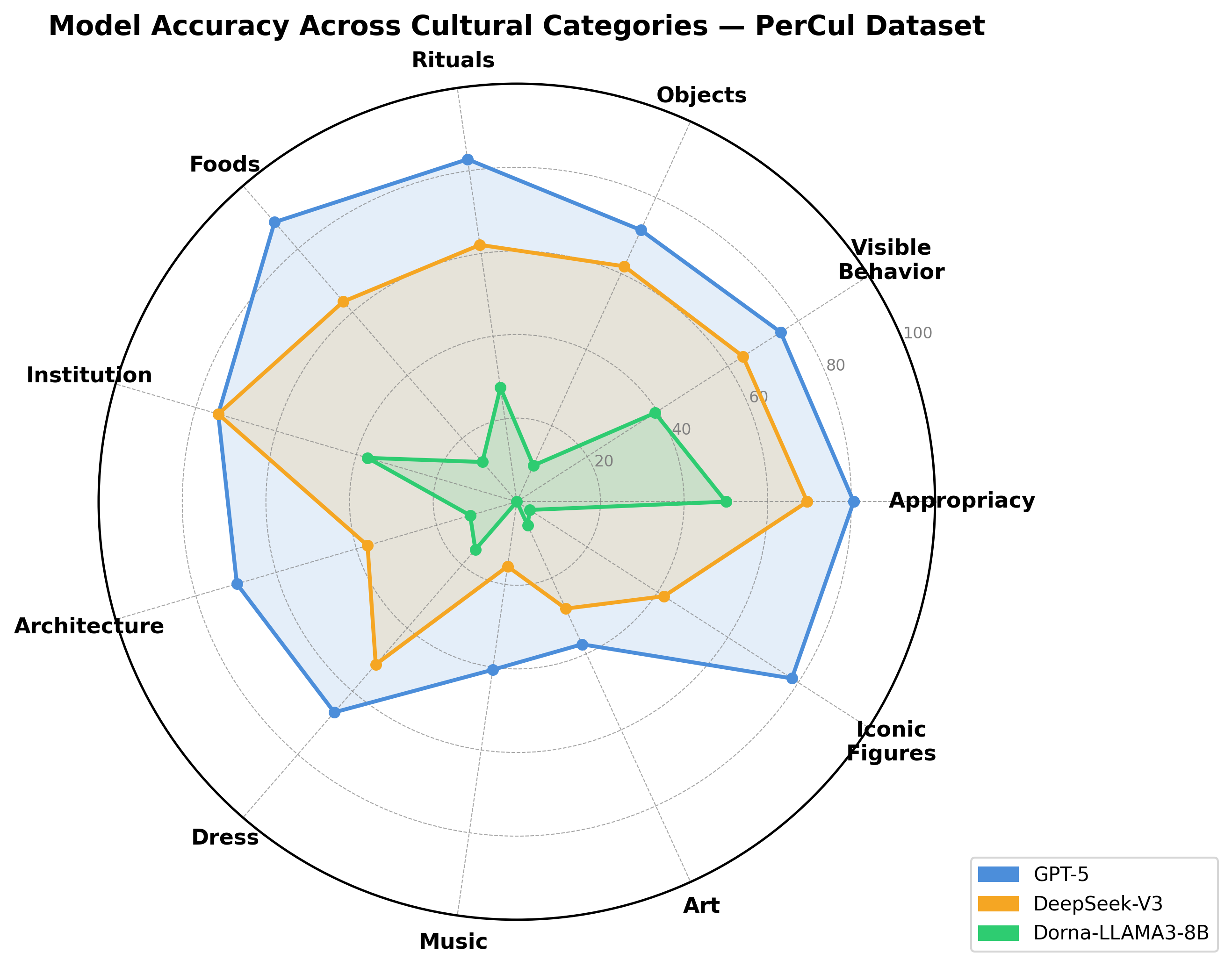}
  \caption{Category-wise accuracy on PerCul-SAQ dataset. The plot shows accuracy for the top three models among each closed-, open, and Persian models. }
  \label{fig:percul-spider}
\end{figure}

Closed-source models perform best across all datasets (Tables~\ref{tab:BLEnD}, \ref{tab:percul}, \ref{tab:isn}). Claude-Opus-4.1 and GPT-5 reach the highest Maux+Post scores, while Gemma-2-27B and DeepSeek-V3 are the strongest open-weight models. PerCul-SAQ and ISN-SAQ are harder than BLEnD, with large drops in lexical scores and semantic metrics. Persian fine-tuned models fall behind: Maral and PersianMind often produce repetitive or meaningless tokens, making their outputs effectively unusable as reflected in zero Rouge scores (See Tables ~\ref{tab:percul}, ~\ref{tab:isn}). Meanwhile, Gemma-2-27B outperforms all Persian-specific models
despite no Persian-targeted fine-tuning. 

\paragraph{Subtopic Analysis}

Based on Figure ~\ref{fig:blend-svg}, we selected the top model across each model family for subtopic analysis of BLEnD. As seen in Figure~\ref{fig:blend-spider}, on BLEnD, Claude Opus performs consistently above 76\% across all subtopics, peaking at Holidays/Celebration/Leisure (89.6\%) and Education (87.8\%), while its weakest category, Family (76.3\%), still surpasses both smaller models by a wide margin. Notably, Ava (8B) and Gemma (27B) exhibit similar performance profiles despite their substantial difference in parameter count, with overlapping scores across nearly every category: Gemma's best is Work Life (65.6\%) and Ava's is Sport (65.4\%). Both models struggle most with Food (Gemma: 45.8\%, Ava: 49.0\%) and Family (Gemma: 50.9\%, Ava: 45.8\%), pointing to persistent gaps in everyday cultural knowledge.

As seen in Figure ~\ref{fig:percul-spider}, performance varies sharply across cultural subtopics on PerCul-SAQ. GPT-5 leads across all categories, with highest accuracy in Foods (88.5\%) and Rituals (82.8\%), and even its weakest categories -- Art (37.5\%) and Music (40.6\%) -- remain competitive.
DeepSeek performs best on Institution (74.4\%) and Appropriacy (69.4\%), but drops sharply in Music (15.6\%) and Art (28.1\%).
Dorna (8B) struggles severely across the board, with near-zero scores in Music (0\%), Iconic Figures (3.6\%), and Art (6.3\%), and only marginally better performance in Appropriacy (50\%) and Visible Behavior (39.3\%), revealing that Persian cultural generalization remains far out of reach for smaller specialized models.

\begin{table}[!ht]
\centering
\resizebox{\columnwidth}{!}{%
\begin{tabular}{@{}lcc@{}}
\toprule
\textbf{Rater Pair} & \textbf{Agreement (\%)} & \textbf{Cohen's $\kappa$} \\
\midrule
LLM-judge vs.\ Annotator 1  & 77.9 & 0.015 \\
LLM-judge vs.\ Annotator 2  & 83.8 & 0.461 \\
LLM-judge vs.\ Annotator 3  & 80.9 & 0.271 \\
\midrule
Annotator 1 vs.\ Annotator 2 & 80.0 & 0.205 \\
Annotator 1 vs.\ Annotator 3 & 85.7 & 0.300 \\
Annotator 2 vs.\ Annotator 3 & 80.0 & 0.300 \\
\midrule
Maux vs.\ Annotator 1       & 93.4 & -0.025 \\
Maux vs.\ Annotator 2       & 85.2 & 0.157 \\
Maux vs.\ Annotator 3       & 90.2 & 0.228 \\
\midrule
LLM-judge vs.\ human majority & 82.4 & 0.244 \\
Maux vs.\ human majority      & 95.1 & 0.384 \\
\bottomrule
\end{tabular}%
}
\caption{Inter-rater agreement between the LLM-judge, Maux, and three human annotators on 70 BLEnD samples. Percent agreement and Cohen's $\kappa$ are reported for all pairwise combinations. Maux achieves higher agreement with human majority than LLM-judge across both metrics.}
\label{tab:human_eval}
\end{table}

\paragraph{Human Evaluation}

Three annotators independently labeled 70 GPT-5 BLEnD responses as correct (1) or incorrect (0) to validate our automatic metrics. Table~\ref{tab:human_eval} reports percent agreement and Cohen's $\kappa$ across all rater pairs. Human annotators agree 80.0--85.7\% of the time with $\kappa$ ranging 0.205--0.300, reflecting moderate agreement given the open-ended nature of the task. Low $\kappa$ values across all pairs -- including among humans -- reflect a known prevalence artifact~\cite{feinstein1990high}: with over 85\% of labels positive, $\kappa$ is deflated by design and raw agreement is the more informative signal.

The LLM-judge achieves 77.9--83.8\% agreement against individual annotators and 82.4\% against the human majority ($\kappa = 0.244$), but is systematically stricter, producing 9 false negatives versus only 3 false positives -- suggesting it penalizes surface-divergent but semantically correct responses. Maux outperforms the LLM-judge on both metrics, achieving 85.2--93.4\% per-annotator agreement and 95.1\% against the human majority ($\kappa = 0.384$). The negative $\kappa$ for Maux vs.\ Annotator~1 ($\kappa = $-$0.025$) is an artifact of near-zero variance in that annotator's labels rather than genuine disagreement. Taken together, these results suggest that Maux better captures human judgment of semantic correctness for Persian cultural responses than the LLM-judge.

\section{Conclusion}
\label{sec:conclusion}

We present TARAZ, a comprehensive evaluation framework for assessing Persian cultural competence in large language models through short-answer questions. Our contribution addresses critical gaps in existing Persian NLP evaluation: while prior benchmarks rely on multiple-choice formats that permit pattern matching without genuine understanding, our framework requires open-ended generation and employs semantic similarity metrics robust to Persian's morphological complexity. Through systematic evaluation of 15 state-of-the-art models, we demonstrate that semantic similarity with Persian-specific postprocessing outperforms traditional lexical metrics. Our Persian-specific postprocessing pipeline handles numeric normalization, conjunction segmentation, and morphological variation which proves essential for robust evaluation, with consistent improvements across all model categories. 

Our findings reveal that closed-source models consistently lead across all three datasets, with GPT-5 and Claude Opus achieving the strongest results on BLEnD and PerCul-SAQ, though all models -- including frontier ones -- struggle considerably on ISN-SAQ, revealing that Iranian social norm reasoning remains an open challenge. Among open-weight models, Gemma-2-27B performs remarkably well on BLEnD (EM: 70.6\%), approaching closed-source levels, while DeepSeek-V3 shows the strongest open-weight performance on PerCul-SAQ and ISN-SAQ. Persian fine-tuned models lag far behind across all three datasets and all metrics, underscoring that language-specific fine-tuning alone is insufficient and that Persian cultural and social reasoning requires substantially more targeted training investment.

We publicly release all datasets, evaluation code, and postprocessing modules to establish a standardized benchmark for Persian cultural understanding and provide a reproducible foundation for cross-cultural LLM evaluation research. 

\section*{Limitations}
\label{sec:limitations}

Our work has several limitations. We focus only on text-based evaluation and do not cover multimodal or vision-language understanding, which is an important direction for future research. Like all human-annotated datasets, ours may reflect annotator bias and fail to capture the full range of cultural interpretations across Persian-speaking regions and demographics.

Additionally, our framework is language-specific and tailored to Persian’s morphology and cultural context. While the general method can be applied to other morphologically rich languages, the normalization and postprocessing steps would need to be redesigned. 

Moreover, both Maux metrics depend on \textit{maux-gte-persian-v3}, which is fine-tuned on formal Persian text and may not favor colloquial text. Bias analysis of this embedding model remains future work.

Finally, the poor performance of Persian fine-tuned models likely stems from three factors: training on machine-translated data introduces translationese artifacts; fine-tuning is predominantly applied to 7--8B parameter models, far below the 27--72B scale of stronger multilingual models; and instruction tuning optimizes format rather than injecting cultural knowledge. Future Persian LLMs should prioritize native instruction data and culturally grounded pretraining over language-specific adaptation alone.

\section*{Ethics Statement}
This work involves human annotation of model outputs for evaluation purposes. Annotators were recruited from the research team and provided informed consent. The datasets used (BLEnD, PerCul, ISN) are publicly available and were collected by their respective authors under appropriate ethical guidelines. Our ISN-SAQ and PerCul-SAQ adaptations involve no new human subjects data collection. The cultural norms described in ISN-SAQ reflect documented Iranian social practices and do not represent the views of the authors. All models were accessed via official APIs or public repositories.

\renewcommand{\refname}{Bibliographical References}
\bibliographystyle{lrec2026-natbib}
\bibliography{lrec2026-example}

@inproceedings{arora2025calmqa,
    title = "{C}a{LMQA}: Exploring culturally specific long-form question answering across 23 languages",
    author = "Arora, Shane  and
      Karpinska, Marzena  and
      Chen, Hung-Ting  and
      Bhattacharjee, Ipsita  and
      Iyyer, Mohit  and
      Choi, Eunsol",
    editor = "Che, Wanxiang  and
      Nabende, Joyce  and
      Shutova, Ekaterina  and
      Pilehvar, Mohammad Taher",
    booktitle = "Proceedings of the 63rd Annual Meeting of the Association for Computational Linguistics (Volume 1: Long Papers)",
    month = jul,
    year = "2025",
    address = "Vienna, Austria",
    publisher = "Association for Computational Linguistics",
    url = "https://aclanthology.org/2025.acl-long.578/",
    doi = "10.18653/v1/2025.acl-long.578",
    pages = "11772--11817",
    ISBN = "979-8-89176-251-0"
}

@inproceedings{shenculturalcommonsens,
    title = "Understanding the Capabilities and Limitations of Large Language Models for Cultural Commonsense",
    author = "Shen, Siqi  and
      Logeswaran, Lajanugen  and
      Lee, Moontae  and
      Lee, Honglak  and
      Poria, Soujanya  and
      Mihalcea, Rada",
    editor = "Duh, Kevin  and
      Gomez, Helena  and
      Bethard, Steven",
    booktitle = "Proceedings of the 2024 Conference of the North American Chapter of the Association for Computational Linguistics: Human Language Technologies (Volume 1: Long Papers)",
    month = jun,
    year = "2024",
    address = "Mexico City, Mexico",
    publisher = "Association for Computational Linguistics",
    url = "https://aclanthology.org/2024.naacl-long.316/",
    doi = "10.18653/v1/2024.naacl-long.316",
    pages = "5668--5680"
}

@inproceedings{horbach2024llms,
    title = "{LLM}s in Short Answer Scoring: Limitations and Promise of Zero-Shot and Few-Shot Approaches",
    author = "Chamieh, Imran  and
      Zesch, Torsten  and
      Giebermann, Klaus",
    editor = {Kochmar, Ekaterina  and
      Bexte, Marie  and
      Burstein, Jill  and
      Horbach, Andrea  and
      Laarmann-Quante, Ronja  and
      Tack, Ana{\"i}s  and
      Yaneva, Victoria  and
      Yuan, Zheng},
    booktitle = "Proceedings of the 19th Workshop on Innovative Use of NLP for Building Educational Applications (BEA 2024)",
    month = jun,
    year = "2024",
    address = "Mexico City, Mexico",
    publisher = "Association for Computational Linguistics",
    url = "https://aclanthology.org/2024.bea-1.25/",
    pages = "309--315"
}

@article{khashabi2021parsinlu,
    title = "{P}arsi{NLU}: A Suite of Language Understanding Challenges for {P}ersian",
    author = "Khashabi, Daniel  and
      Cohan, Arman  and
      Shakeri, Siamak  and
      Hosseini, Pedram and others",
    editor = "Roark, Brian  and
      Nenkova, Ani",
    journal = "Transactions of the Association for Computational Linguistics",
    volume = "9",
    year = "2021",
    address = "Cambridge, MA",
    publisher = "MIT Press",
    url = "https://aclanthology.org/2021.tacl-1.68/",
    doi = "10.1162/tacl_a_00419",
    pages = "1147--1162"
}

@inproceedings{moosavi2025percul,
    title = "{P}er{C}ul: A Story-Driven Cultural Evaluation of {LLM}s in {P}ersian",
    author = "Moosavi Monazzah, Erfan  and
      Rahimzadeh, Vahid  and
      Yaghoobzadeh, Yadollah  and
      Shakery, Azadeh  and
      Pilehvar, Mohammad Taher",
    editor = "Chiruzzo, Luis  and
      Ritter, Alan  and
      Wang, Lu",
    booktitle = "Proceedings of the 2025 Conference of the Nations of the Americas Chapter of the Association for Computational Linguistics: Human Language Technologies (Volume 1: Long Papers)",
    month = apr,
    year = "2025",
    address = "Albuquerque, New Mexico",
    publisher = "Association for Computational Linguistics",
    url = "https://aclanthology.org/2025.naacl-long.631/",
    doi = "10.18653/v1/2025.naacl-long.631",
    pages = "12670--12687",
    ISBN = "979-8-89176-189-6"
}

@misc{myung2024blend,
      title={BLEnD: A Benchmark for LLMs on Everyday Knowledge in Diverse Cultures and Languages}, 
      author={Junho Myung and Nayeon Lee and Yi Zhou and Jiho Jin and Rifki Afina Putri and Dimosthenis Antypas and Hsuvas Borkakoty and Eunsu Kim and Carla Perez-Almendros and others},
      year={2024},
      eprint={2406.09948},
      archivePrefix={arXiv},
      primaryClass={cs.CL},
      url={https://arxiv.org/abs/2406.09948}, 
}

@inproceedings{sadr2025taarofbench,
    title = "We Politely Insist: Your {LLM} Must Learn the {P}ersian Art of Taarof",
    author = "Sadr, Nikta Gohari  and
      Heidariasl, Sahar  and
      Megerdoomian, Karine  and
      Seyyed-Kalantari, Laleh  and
      Emami, Ali",
    editor = "Christodoulopoulos, Christos  and
      Chakraborty, Tanmoy  and
      Rose, Carolyn  and
      Peng, Violet",
    booktitle = "Proceedings of the 2025 Conference on Empirical Methods in Natural Language Processing",
    month = nov,
    year = "2025",
    address = "Suzhou, China",
    publisher = "Association for Computational Linguistics",
    url = "https://aclanthology.org/2025.emnlp-main.94/",
    doi = "10.18653/v1/2025.emnlp-main.94",
    pages = "1819--1838",
    ISBN = "979-8-89176-332-6"
}

@inproceedings{saffari2025iranian,
    title = "Can {I} Introduce My Boyfriend to My Grandmother? Evaluating Large Language Models Capabilities on {I}ranian Social Norm Classification",
    author = "Saffari, Hamidreza  and
      Shafiei, Mohammadamin  and
      Rooein, Donya  and
      Pierri, Francesco  and
      Nozza, Debora",
    editor = "Chiruzzo, Luis  and
      Ritter, Alan  and
      Wang, Lu",
    booktitle = "Findings of the Association for Computational Linguistics: NAACL 2025",
    month = apr,
    year = "2025",
    address = "Albuquerque, New Mexico",
    publisher = "Association for Computational Linguistics",
    url = "https://aclanthology.org/2025.findings-naacl.337/",
    doi = "10.18653/v1/2025.findings-naacl.337",
    pages = "6060--6074",
    ISBN = "979-8-89176-195-7"
}

@inproceedings{brown2020language,
 author = {Brown, Tom and Mann, Benjamin and Ryder, Nick and Subbiah, Melanie and others},
 booktitle = {Advances in Neural Information Processing Systems},
 editor = {H. Larochelle and M. Ranzato and R. Hadsell and M.F. Balcan and H. Lin},
 pages = {1877--1901},
 publisher = {Curran Associates, Inc.},
 title = {Language Models are Few-Shot Learners},
 url = {https://proceedings.neurips.cc/paper_files/paper/2020/file/1457c0d6bfcb4967418bfb8ac142f64a-Paper.pdf},
 volume = {33},
 year = {2020}
}

@misc{shamsfard2025farseval,
      title={FarsEval-PKBETS: A new diverse benchmark for evaluating Persian large language models}, 
      author={Seyed Mohammad Hossein Hashemi and Zahra Vatankhah and Motahareh Ramezani and Niki Pourazin and Tara Zare and others},
      year={2025},
      eprint={2504.14690},
      archivePrefix={arXiv},
      primaryClass={cs.CL},
      url={https://arxiv.org/abs/2504.14690}, 
}

@misc{luo2023,
      title={ChatGPT as a Factual Inconsistency Evaluator for Text Summarization}, 
      author={Zheheng Luo and Qianqian Xie and Sophia Ananiadou},
      year={2023},
      eprint={2303.15621},
      archivePrefix={arXiv},
      primaryClass={cs.CL},
      url={https://arxiv.org/abs/2303.15621}, 
}

@inproceedings{ying2025disentangling,
    title = "Disentangling Language and Culture for Evaluating Multilingual Large Language Models",
    author = "Ying, Jiahao  and
      Tang, Wei  and
      Zhao, Yiran  and
      Cao, Yixin  and
      Rong, Yu  and
      Zhang, Wenxuan",
    editor = "Che, Wanxiang  and
      Nabende, Joyce  and
      Shutova, Ekaterina  and
      Pilehvar, Mohammad Taher",
    booktitle = "Proceedings of the 63rd Annual Meeting of the Association for Computational Linguistics (Volume 1: Long Papers)",
    month = jul,
    year = "2025",
    address = "Vienna, Austria",
    publisher = "Association for Computational Linguistics",
    url = "https://aclanthology.org/2025.acl-long.1082/",
    doi = "10.18653/v1/2025.acl-long.1082",
    pages = "22230--22251",
    ISBN = "979-8-89176-251-0"
}

@misc{abbasi2023persianllama,
      title={PersianLLaMA: Towards Building First Persian Large Language Model}, 
      author={Mohammad Amin Abbasi and Arash Ghafouri and Mahdi Firouzmandi and Hassan Naderi and Behrouz Minaei Bidgoli},
      year={2023},
      eprint={2312.15713},
      archivePrefix={arXiv},
      primaryClass={cs.CL},
      url={https://arxiv.org/abs/2312.15713}, 
}

@misc{qwen2025qwen25,
      title={Qwen2.5 Technical Report}, 
      author={Qwen and Yang, An and others},
      year={2025},
      eprint={2412.15115},
      archivePrefix={arXiv},
      primaryClass={cs.CL},
      url={https://arxiv.org/abs/2412.15115}
}

@misc{touvron2023llama,
      title={LLaMA: Open and Efficient Foundation Language Models}, 
      author={Hugo Touvron and Thibaut Lavril and Gautier Izacard and Xavier Martinet and Marie-Anne Lachaux and Timothée Lacroix and Baptiste Rozière and Naman Goyal and Eric Hambro and Faisal Azhar and Aurelien Rodriguez and Armand Joulin and Edouard Grave and Guillaume Lample},
      year={2023},
      eprint={2302.13971},
      archivePrefix={arXiv},
      primaryClass={cs.CL},
      url={https://arxiv.org/abs/2302.13971}, 
}

@inproceedings{zhang-etal-2024-mgte,
    title = "{mGTE}: Generalized Long-Context Text Representation and Reranking Models for Multilingual Text Retrieval",
    author = "Zhang, Xin  and
      Zhang, Yanzhao  and
      Long, Dingkun  and
      Xie, Wen  and
      Dai, Ziqi  and
      Tang, Jialong  and
      Lin, Huan  and
      others",
    editor = "Dernoncourt, Franck  and
      Preo{\c{t}}iuc-Pietro, Daniel  and
      Shimorina, Anastasia",
    booktitle = "Proceedings of the 2024 Conference on Empirical Methods in Natural Language Processing: Industry Track",
    month = nov,
    year = "2024",
    address = "Miami, Florida, US",
    publisher = "Association for Computational Linguistics",
    url = "https://aclanthology.org/2024.emnlp-industry.103/",
    doi = "10.18653/v1/2024.emnlp-industry.103",
    pages = "1393--1412"
}

@inproceedings{lin2004rouge,
    title = "{ROUGE}: A Package for Automatic Evaluation of Summaries",
    author = "Lin, Chin-Yew",
    booktitle = "Text Summarization Branches Out",
    month = jul,
    year = "2004",
    address = "Barcelona, Spain",
    publisher = "Association for Computational Linguistics",
    url = "https://aclanthology.org/W04-1013/",
    pages = "74--81"
}

@misc{li2024culturepark,
      title={CulturePark: Boosting Cross-cultural Understanding in Large Language Models}, 
      author={Cheng Li and Damien Teney and Linyi Yang and Qingsong Wen and Xing Xie and Jindong Wang},
      year={2024},
      eprint={2405.15145},
      archivePrefix={arXiv},
      primaryClass={cs.AI},
      url={https://arxiv.org/abs/2405.15145}, 
}

@misc{ghahroodi2024khayyam,
    title={Khayyam Challenge (PersianMMLU): Is Your LLM Truly Wise to The Persian Language?},
    author={Omid Ghahroodi and Marzia Nouri and Mohammad Vali Sanian and Alireza Sahebi and Doratossadat Dastgheib and Ehsaneddin Asgari and Mahdieh Soleymani Baghshah and Mohammad Hossein Rohban},
    year={2024},
    eprint={2404.06644},
    archivePrefix={arXiv},
    primaryClass={cs.CL}
}

@inproceedings{hosseinbeigi2025advancing,
    title = "Advancing {P}ersian {LLM} Evaluation",
    author = "Hosseinbeigi, Sara Bourbour  and
      Rohani, Behnam  and
      Masoudi, Mostafa  and
      Shamsfard, Mehrnoush  and
      Saaberi, Zahra  and
      Manesh, Mostafa Karimi  and
      Abbasi, Mohammad Amin",
    editor = "Chiruzzo, Luis  and
      Ritter, Alan  and
      Wang, Lu",
    booktitle = "Findings of the Association for Computational Linguistics: NAACL 2025",
    month = apr,
    year = "2025",
    address = "Albuquerque, New Mexico",
    publisher = "Association for Computational Linguistics",
    url = "https://aclanthology.org/2025.findings-naacl.147/",
    doi = "10.18653/v1/2025.findings-naacl.147",
    pages = "2711--2727",
    ISBN = "979-8-89176-195-7"
}

@misc{zhang2025culfit,
      title={CulFiT: A Fine-grained Cultural-aware LLM Training Paradigm via Multilingual Critique Data Synthesis}, 
      author={Ruixiang Feng and Shen Gao and Xiuying Chen and Lisi Chen and Shuo Shang},
      year={2025},
      eprint={2505.19484},
      archivePrefix={arXiv},
      primaryClass={cs.CL},
      url={https://arxiv.org/abs/2505.19484}, 
}

@article{farahani2021parsbert,
   title={ParsBERT: Transformer-based Model for Persian Language Understanding},
   volume={53},
   ISSN={1573-773X},
   url={http://dx.doi.org/10.1007/s11063-021-10528-4},
   DOI={10.1007/s11063-021-10528-4},
   number={6},
   journal={Neural Processing Letters},
   publisher={Springer Science and Business Media LLC},
   author={Farahani, Mehrdad and Gharachorloo, Mohammad and Farahani, Marzieh and Manthouri, Mohammad},
   year={2021},
   month=oct, pages={3831–3847}
}

@misc{rostami2024persianmind,
      title={PersianMind: A Cross-Lingual Persian-English Large Language Model}, 
      author={Pedram Rostami and Ali Salemi and Mohammad Javad Dousti},
      year={2024},
      eprint={2401.06466},
      archivePrefix={arXiv},
      primaryClass={cs.CL},
      url={https://arxiv.org/abs/2401.06466}, 
}

@inproceedings{li2024culturellm,
 author = {Li, Cheng and Chen, Mengzhuo and Wang, Jindong and Sitaram, Sunayana and Xie, Xing},
 booktitle = {Advances in Neural Information Processing Systems},
 editor = {A. Globerson and L. Mackey and D. Belgrave and A. Fan and U. Paquet and J. Tomczak and C. Zhang},
 pages = {84799--84838},
 publisher = {Curran Associates, Inc.},
 title = {CultureLLM: Incorporating Cultural Differences into Large Language Models},
 url = {https://proceedings.neurips.cc/paper_files/paper/2024/file/9a16935bf54c4af233e25d998b7f4a2c-Paper-Conference.pdf},
 volume = {37},
 year = {2024}
}

@inproceedings{matina2025,
    title = "Matina: A Large-Scale 73{B} Token {P}ersian Text Corpus",
    author = "Hosseinbeigi, Sara Bourbour  and
      Taherinezhad, Fatemeh  and
      Faili, Heshaam  and
      Baghbani, Hamed  and
      Nadi, Fatemeh  and
      Amiri, Mostafa",
    editor = "Chiruzzo, Luis  and
      Ritter, Alan  and
      Wang, Lu",
    booktitle = "Proceedings of the 2025 Conference of the Nations of the Americas Chapter of the Association for Computational Linguistics: Human Language Technologies (Volume 1: Long Papers)",
    month = apr,
    year = "2025",
    address = "Albuquerque, New Mexico",
    publisher = "Association for Computational Linguistics",
    url = "https://aclanthology.org/2025.naacl-long.462/",
    doi = "10.18653/v1/2025.naacl-long.462",
    pages = "9143--9157",
    ISBN = "979-8-89176-189-6"
}

@misc{parst5,
  author = {Puraminy, Ahmad},
  title = {{P}ars{T}5: A {T}5 Model for {P}ersian},
  year = {2021},
  howpublished = {\url{https://huggingface.co/Ahmad/parsT5-base}}
}

@misc{dorna2024,
  author = {{PartAI}},
  title = {Dorna-{LL}ama3-8{B}-Instruct},
  year = {2024},
  howpublished = {\url{https://huggingface.co/PartAI/Dorna-Llama3-8B-Instruct}}
}

@misc{maralgpt2023,
  author = {Haghiri, Muhammadreza and Mohrechi, Mahi },
  title = {Maral-7{B}-alpha-1: A {P}ersian Large Language Model},
  year = {2023},
  howpublished = {\url{https://huggingface.co/MaralGPT/Maral-7B-alpha-1}}
}

@misc{moghadam2024ava,
  author = {Moghadam, Mehdi Hosseini},
  title = {{AVA}-{LL}ama-3: Fine-Tuned {LL}ama 3 {P}ersian Large Language Model},
  year = {2024},
  howpublished = {\url{https://github.com/mehdihosseinimoghadam/AVA-Llama-3}}
}

@misc{openai2025gpt5,
  author = {{OpenAI}},
  title = {Introducing {GPT}-5},
  year = {2025},
  howpublished = {\url{https://openai.com/index/introducing-gpt-5/}}
}

@techreport{anthropic2025claude4,
  author    = {Anthropic},
  title     = {System Card Addendum: {Claude Opus 4.1}},
  year      = {2025},
  month     = {August},
  url       = {https://www-cdn.anthropic.com/9fa30625273bafdf5af82c93719d7ca606485a16.pdf}
}

@techreport{anthropic2025sonnet45,
  author    = {Anthropic},
  title     = {System Card: {Claude Sonnet 4.5}},
  year      = {2025},
  month     = {September},
  url       = {https://assets.anthropic.com/m/12f214efcc2f457a/original/Claude-Sonnet-4-5-System-Card.pdf}
}

@misc{google2025gemini25,
      title={Gemini 2.5: Pushing the Frontier with Advanced Reasoning, Multimodality, Long Context, and Next Generation Agentic Capabilities}, 
      author={Comanici, Gheorghe and others},
      year={2025},
      eprint={2507.06261},
      archivePrefix={arXiv},
      primaryClass={cs.CL},
      url={https://arxiv.org/abs/2507.06261}
}

@misc{deepseek2024,
  title={DeepSeek-V3 Technical Report},
  author={DeepSeek-AI and Liu, Aixin and others},
  year={2025},
  eprint={2412.19437},
  archivePrefix={arXiv},
  primaryClass={cs.CL},
  url={https://arxiv.org/abs/2412.19437}
}

@misc{team2025gemma,
      title={Gemma 2: Improving Open Language Models at a Practical Size}, 
      author={Gemma Team and Riviere, Morgane and others},
      year={2024},
      eprint={2408.00118},
      archivePrefix={arXiv},
      primaryClass={cs.CL},
      url={https://arxiv.org/abs/2408.00118}
}

@misc{microsoft2025phi4,
      title={Phi-4-reasoning Technical Report}, 
      author={Abdin, Marah and others},
      year={2025},
      eprint={2504.21318},
      archivePrefix={arXiv},
      primaryClass={cs.AI},
      url={https://arxiv.org/abs/2504.21318}
}

@misc{shariati2025parsbench,
  author = {Shariati Motlagh, Shahriar},
  title = {{P}ars{B}ench: A Toolkit for Benchmarking {P}ersian Language Models},
  year = {2025},
  howpublished = {\url{https://github.com/ParsBench/ParsBench}},
  note = {GitHub repository}
}

@misc{mizan,
  author       = {{MCINext Team}},
  title        = {MIZAN: A Persian LLM Leaderboard},
  year         = {2025},
  howpublished = {Hamrah Aval (MCI Next), hosted on Hugging Face},
  url          = {https://huggingface.co/spaces/MCINext/mizan-llm-leaderboard},
  publisher    = {Hugging Face}
}

@misc{PartAI_OpenPersianLLM_2024,
  author       = {{PartAI and AUT NLP Lab}},
  title        = {Open Persian LLM Leaderboard},
  year         = {2024},
  howpublished = {Developed by PartAI in collaboration with Amirkabir University of Technology (AUT) NLP Lab},
  url          = {https://huggingface.co/spaces/PartAI/open-persian-llm-leaderboard},
  publisher    = {Hugging Face}
}

@inproceedings{goyal2025ember,
    title = "Are {LLM}-Judges Robust to Expressions of Uncertainty? Investigating the effect of Epistemic Markers on {LLM}-based Evaluation",
    author = "Lee, Dongryeol  and
      Hwang, Yerin  and
      Kim, Yongil  and
      Park, Joonsuk  and
      Jung, Kyomin",
    editor = "Chiruzzo, Luis  and
      Ritter, Alan  and
      Wang, Lu",
    booktitle = "Proceedings of the 2025 Conference of the Nations of the Americas Chapter of the Association for Computational Linguistics: Human Language Technologies (Volume 1: Long Papers)",
    month = apr,
    year = "2025",
    address = "Albuquerque, New Mexico",
    publisher = "Association for Computational Linguistics",
    url = "https://aclanthology.org/2025.naacl-long.452/",
    doi = "10.18653/v1/2025.naacl-long.452",
    pages = "8962--8984",
    ISBN = "979-8-89176-189-6"
}

@misc{mauxgte2024,
  author = {xmanii},
  title = {maux-gte-persian-v3: High-performance {P}ersian sentence embedding model},
  year = {2024},
  howpublished = {\url{https://huggingface.co/xmanii/maux-gte-persian-v3}}
}

@inproceedings{roitman2024exam,
  title = {EXAM++: LLM-based Answerability Metrics for IR Evaluation},
  author = {Naghmeh Farzi and Laura Dietz},
  booktitle = {LLM4Eval: The First Workshop on Large Language Models for Evaluation in Information Retrieval},
  address = {Washington DC, United States},
  year = {2024},
  pages = {1--20},
  url = {https://www.cs.unh.edu/~dietz/papers/farzi2024exampp.pdf},
  organization = {CEUR Workshop Proceedings}
}

@article{feinstein1990high,
  author  = {Feinstein, Alvan R. and Cicchetti, Domenic V.},
  title   = {High agreement but low kappa: {I}. {T}he problems of two paradoxes},
  journal = {Journal of Clinical Epidemiology},
  year    = {1990},
  volume  = {43},
  number  = {6},
  pages   = {543-549},
}
\label{sec:reference}

\section*{Appendix A. Dataset Examples}
\label{app:example}

\begin{figure}[h]
\centering
\begin{mdframed}[
  backgroundcolor=boxbg,
  linecolor=boxborder,
  linewidth=1.5pt,
  roundcorner=6pt,
  innerleftmargin=14pt,
  innerrightmargin=14pt,
  innertopmargin=12pt,
  innerbottommargin=12pt
]
{\small\textcolor{labelgray}{\texttt{ID:}}\;\textbf{\texttt{Al-en-01}}\hfill
\textcolor{goldtag}{\textbf{Topic:}}\;\textcolor{goldtag}{Food}}
\vspace{6pt}\hrule\vspace{8pt}

\begin{mdframed}[
  backgroundcolor=promptbg,
  linecolor=promptborder,
  linewidth=1pt,
  topline=false, rightline=false, bottomline=false,
  innerleftmargin=10pt, innerrightmargin=6pt,
  innertopmargin=6pt, innerbottommargin=6pt
]
{\footnotesize\textcolor{labelgray}{\textsc{Question}}}\\[4pt]
\begin{RTL}\begin{persian}\setRL
{\small یک میان وعده معمول برای بچه‌های پیش دبستانی در ایران چیست؟}
\end{persian}\end{RTL}

\vspace{2pt}
{\footnotesize\textit{(What is a common snack for preschool kids in Iran?)}}
\end{mdframed}

\vspace{6pt}

\begin{mdframed}[
  backgroundcolor=answerbg,
  linecolor=answerborder,
  linewidth=1pt,
  roundcorner=4pt,
  innerleftmargin=10pt, innerrightmargin=10pt,
  innertopmargin=6pt, innerbottommargin=6pt
]
{\footnotesize\textcolor{labelgray}{\textsc{Accepted Answers (with annotator frequency)}}}\\[6pt]
\begin{tabular}{@{}lll@{}}
\textbf{Persian} & \textbf{English} & \textbf{Count} \\
\midrule
\begin{RTL}\begin{persian}\setRL میوه / ميوه\end{persian}\end{RTL} & fruit & 3 \\
\begin{RTL}\begin{persian}\setRL لقمه\end{persian}\end{RTL} & sandwich & 1 \\
\begin{RTL}\begin{persian}\setRL شیر و کیک\end{persian}\end{RTL} & cake and milk & 1 \\
\begin{RTL}\begin{persian}\setRL پنیر و نون\end{persian}\end{RTL} & bread and cheese & 1 \\
\begin{RTL}\begin{persian}\setRL تخم‌مرغ\end{persian}\end{RTL} & eggs & 1 \\
\begin{RTL}\begin{persian}\setRL کورنفلکس\end{persian}\end{RTL} & cereal & 1 \\
\end{tabular}
\end{mdframed}

\end{mdframed}
\caption{An example from the BLEnD Persian subset. Multiple valid answers with annotator frequency counts reflect genuine variation in Iranian households -- fruit (\textfarsi{میوه}) is the most common response, while bread and cheese (\textfarsi{نون و پنير}), a staple of Iranian home meals, also appears. This diversity motivates weighted evaluation beyond exact match.}
\label{fig:blend-example}
\end{figure}

\begin{center}
\begin{mdframed}[
  backgroundcolor=boxbg,
  linecolor=boxborder,
  linewidth=1.5pt,
  roundcorner=6pt,
  innerleftmargin=14pt,
  innerrightmargin=14pt,
  innertopmargin=12pt,
  innerbottommargin=12pt
]

{\small\textcolor{labelgray}{\texttt{ID:}}\;\textbf{\texttt{1\_O\_MUS}}\hfill
\textcolor{goldtag}{\textbf{Category:}}\;\textcolor{goldtag}{\textfarsi{ساز یا موسیقی}}}

\vspace{6pt}\hrule\vspace{8pt}

{\footnotesize\textcolor{labelgray}{\textsc{Story}}}

\begin{RTL}
\begin{persian}
\setRL
{\small
در یک شب بهاری، در یکی از روستاهای قدیمی ایران، مردم برای جشن نوروز گرد هم آمده بودند.
صدای موسیقی از دور به گوش می‌رسید و همه با شور و شوق به سمت میدان اصلی روستا حرکت می‌کردند.
در مرکز میدان، گروهی از نوازندگان با لباس‌های رنگارنگ و سنتی در حال نواختن بودند.
یکی از نوازندگان با مهارت خاصی بر روی یک ساز کوبه‌ای می‌نواخت که صدای آن به خوبی با دیگر سازها
هماهنگ بود و ریتمی شاد و پرانرژی به موسیقی می‌بخشید. مردم با شنیدن این صدا به وجد آمده و
شروع به رقص و پایکوبی کردند. این ساز کوبه‌ای که در دست نوازنده بود، به نوعی روح جشن را
به نمایش می‌گذاشت و همه را به حرکت وادار می‌کرد.
}
\end{persian}
\end{RTL}

\vspace{8pt}

\begin{mdframed}[ backgroundcolor=promptbg,
  linecolor=promptborder,
  linewidth=1pt,
  topline=false, rightline=false, bottomline=false,
  innerleftmargin=10pt,
  innerrightmargin=6pt,
  innertopmargin=6pt,
  innerbottommargin=6pt
]
{\footnotesize\textcolor{labelgray}{\textsc{Prompt (New)}}}\\[4pt]
\begin{RTL}
\begin{persian}
\setRL
{\small
با توجه به متن بالا چه ساز یا موسیقی در فرهنگ یا زندگی روزمره ایرانی قابل برداشت است؟

پاسخ خود را بدون هیچ توضیحی ارائه دهید.
}
\end{persian}
\end{RTL}
\end{mdframed}

\vspace{6pt}

\begin{mdframed}[
  backgroundcolor=answerbg,
  linecolor=answerborder,
  linewidth=1pt,
  roundcorner=4pt,
  innerleftmargin=10pt,
  innerrightmargin=10pt,
  innertopmargin=6pt,
  innerbottommargin=6pt
]
\begin{tabular}{@{}ll@{}}
  {\footnotesize\textcolor{labelgray}{\textsc{Ground Truth}}} &
  {\footnotesize\textcolor{labelgray}{\textsc{GPT-5 Response}}} \\[3pt]
  \textbf{\textcolor{answerborder}{\large\textfarsi{تنبک}}} \;\checkmark &
  \textbf{\textcolor{answerborder}{\large\textfarsi{دف}}} \;\xmark \\
\end{tabular}
\end{mdframed}

\end{mdframed}
\captionof{figure}{An example from the PerCul-SAQ dataset (entry \texttt{1\_O\_MUS}, Music category).
The story implicitly encodes a Persian cultural artifact; GPT-5 incorrectly identifies \textfarsi{دف} (Daf) as the answer.}
\label{fig:percul-example}
\end{center}

\begin{center}
\begin{mdframed}[
  backgroundcolor=boxbg,
  linecolor=boxborder,
  linewidth=1.5pt,
  roundcorner=6pt,
  innerleftmargin=14pt,
  innerrightmargin=14pt,
  innertopmargin=12pt,
  innerbottommargin=12pt
]

{\small\textcolor{labelgray}{\texttt{ID:}}\;\textbf{\texttt{73}}\hfill
\textcolor{goldtag}{\textbf{Label:}}\;\textcolor{answerborder}{\textbf{Normal}}\hfill
\textcolor{goldtag}{\textbf{Scope:}}\;\textcolor{labelgray}{Specific}}

\vspace{6pt}\hrule\vspace{8pt}

{\footnotesize\textcolor{labelgray}{\textsc{Environment}}}

\vspace{4pt}
{\small \textfarsi{بازار} \quad $\rightarrow$ \quad Bazaar}

\vspace{8pt}

\begin{mdframed}[
  backgroundcolor=promptbg,
  linecolor=promptborder,
  linewidth=1pt,
  topline=false, rightline=false, bottomline=false,
  innerleftmargin=10pt,
  innerrightmargin=6pt,
  innertopmargin=6pt,
  innerbottommargin=6pt
]
{\footnotesize\textcolor{labelgray}{\textsc{Original Dataset — Norm \& Context}}}

\vspace{4pt}
\begin{RTL}
\begin{persian}
\setRL
\begin{tabular}{@{}ll@{}}
& {\small چانه‌زنی و مذاکره بر سر قیمت‌ها با فروشندگان}   FA: \\
\end{tabular}
\end{persian}
\end{RTL}

\vspace{4pt}
\begin{tabular}{@{}ll@{}}
  EN: & {\small Bargaining and negotiating} \\
               & {\small prices with vendors} \\
\end{tabular}
\end{mdframed}

\vspace{4pt}
{\footnotesize\textcolor{labelgray}{$\downarrow$\quad We rewrote each original norm into a Short Answer Question (SAQ).}}
\vspace{4pt}

\begin{mdframed}[
  backgroundcolor=answerbg,
  linecolor=answerborder,
  linewidth=1pt,
  topline=false, rightline=false, bottomline=false,
  innerleftmargin=10pt,
  innerrightmargin=6pt,
  innertopmargin=6pt,
  innerbottommargin=6pt
]
{\footnotesize\textcolor{labelgray}{\textsc{Rewritten SAQ Prompt}}}

\begin{RTL}
\begin{persian}
\setRL
{\small در بازار ایران هنگام خرید با فروشندگان چه رفتاری رایج است؟}
\end{persian}
\end{RTL}
\end{mdframed}

\vspace{6pt}

\begin{mdframed}[
  backgroundcolor=answerbg,
  linecolor=answerborder,
  linewidth=1pt,
  roundcorner=4pt,
  innerleftmargin=10pt,
  innerrightmargin=10pt,
  innertopmargin=6pt,
  innerbottommargin=6pt
]
\begin{tabular}{@{}ll@{}}
  {\footnotesize\textcolor{labelgray}{\textsc{Ground Truth}}} &
  {\footnotesize\textcolor{labelgray}{\textsc{GPT-5 Response}}} \\[3pt]
  \begin{RTL}\begin{persian}\setRL\textbf{\textcolor{answerborder}{\large چانه‌زنی}}\end{persian}\end{RTL}\;\checkmark &
  \begin{RTL}\begin{persian}\setRL\textbf{\textcolor{answerborder}{\large چانه‌زنی بر سر قیمت}}\end{persian}\end{RTL}\;\checkmark \\
\end{tabular}
\end{mdframed}

\end{mdframed}
\captionof{figure}{An example from the ISN-SAQ dataset (entry \texttt{73}). The original norm is shown as-is
from the dataset; we rewrote it into a Short Answer Question (SAQ) for evaluation. The English translation is: "What behavior is common when shopping with vendors in Iranian bazaars?"
GPT-5 correctly identifies \textfarsi{چانه‌زنی} (bargaining) as the expected social norm.}
\label{fig:psn-example}
\end{center}

\end{document}